%% file: main.tex
\documentclass[10pt,twocolumn,letterpaper]{article}

\usepackage{iccv}              % To produce the CAMERA-READY version
\input{preamble}

\definecolor{iccvblue}{rgb}{0.21,0.49,0.74}
\usepackage[pagebackref,breaklinks,colorlinks,allcolors=iccvblue]{hyperref}
\usepackage{times}
\usepackage{epsfig}
\usepackage{graphicx}
\usepackage{amsmath}
\usepackage{amssymb}
\usepackage{booktabs}
\usepackage{algorithm}
\usepackage{algpseudocode}
\usepackage{xcolor}
\usepackage{caption}
\usepackage{enumerate}
\usepackage{enumitem}
\usepackage{multirow}
\usepackage[accsupp]{axessibility}  % Improves PDF readability for those with disabilities.

\def\paperID{3021} % *** Enter the Paper ID here
\def\confName{ICCV}
\def\confYear{2025}

\title{Nautilus: Locality-aware Autoencoder for Scalable Mesh Generation}

\author{
\textbf{Yuxuan Wang}\textsuperscript{1$*$}, \textbf{Xuanyu Yi}\textsuperscript{1$*$}, \textbf{Haohan Weng}\textsuperscript{2,3$*$}, \textbf{Qingshan Xu}\textsuperscript{1}, \textbf{Xiaokang Wei}\textsuperscript{4},\\
\textbf{Xianghui Yang}\textsuperscript{2}, \textbf{Chunchao Guo}\textsuperscript{2}, \textbf{Long Chen}\textsuperscript{5},
\textbf{Hanwang Zhang}\textsuperscript{1}\\
{\small \textsuperscript{1}Nanyang Technological University, \textsuperscript{2}Tencent Hunyuan, \textsuperscript{3}South China University of Technology} \\
{\small \textsuperscript{4}The Hong Kong Polytechnic University, \textsuperscript{5}Hong Kong University of Science and Technology}
}

\begin{document}

\twocolumn[{
\renewcommand\twocolumn[1][]{#1}
\maketitle
\centering
\vspace{-0.5cm}
\includegraphics[width=174mm]{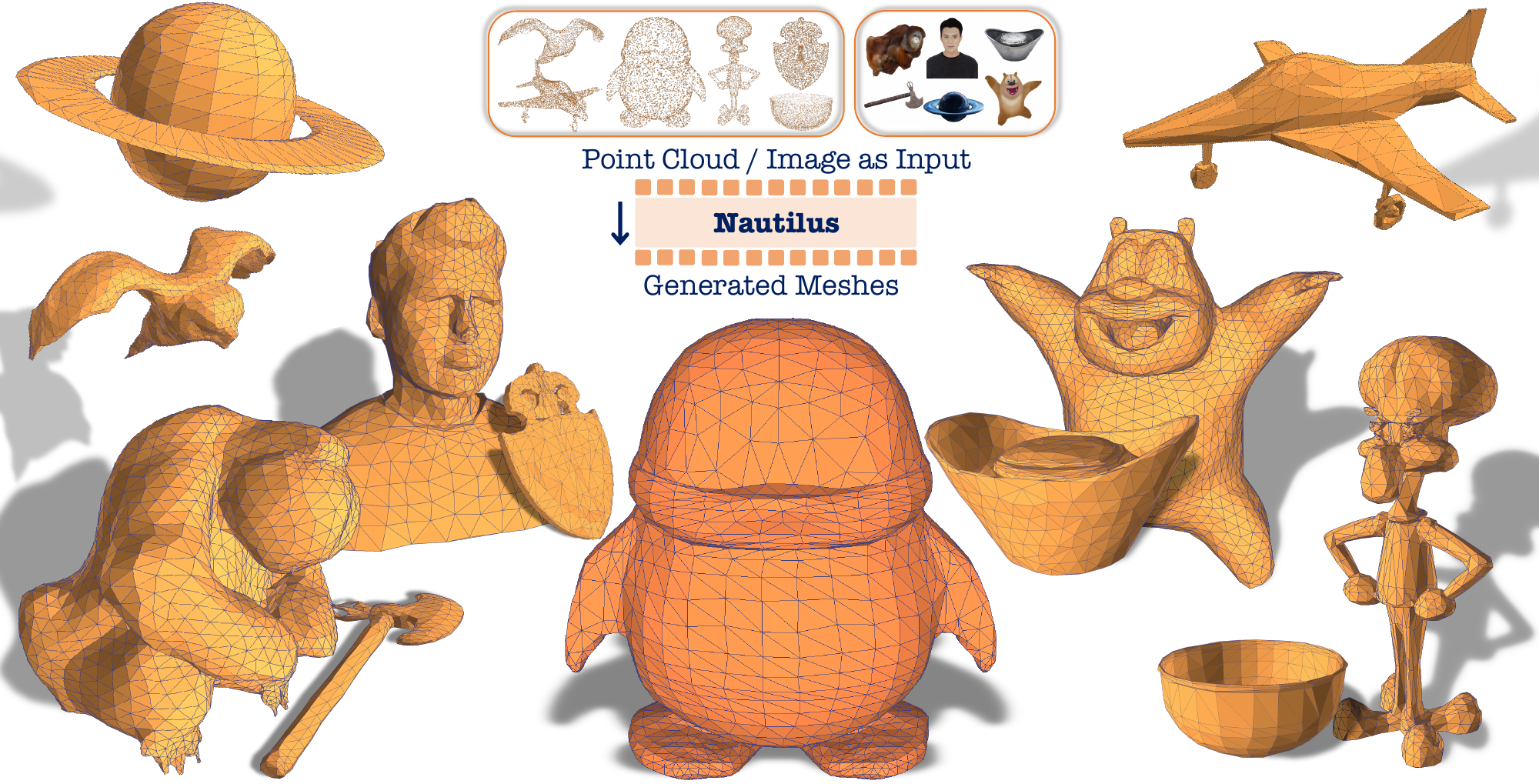} 
\captionsetup{type=figure}
\vspace{-0.5cm}
\caption{Given a point cloud (e.g., the \textit{bird}, \textit{aircraft}, \textit{penguin}, \textit{Squog}, \textit{shield}, and \textit{bowl} shown in the figure) or a single image (e.g., the \textit{gorilla}, \textit{ax}, \textit{man}, \textit{planet}, \textit{sycee}, and \textit{bear} shown in the figure) as input, \textbf{Nautilus} enables the direct generation of mesh assets.
% By explicitly modeling and leveraging locality, it delivers superior generation quality for unprecedentedly complex and detailed topologies.
}
\vspace{0.3cm}
\label{fig:1}
}]

\maketitle

\input{sec/0_abstract}    
\input{sec/1_intro}
\input{sec/2_related_work}

\input{sec/3_method}
\input{sec/4_experiment}
\input{sec/5_conclusion}

\section*{Acknowledgments}
This research is supported by the National Research Foundation, Singapore under its AI Singapore Programme (AISG Award No: AISG3-RP-2022-030).

{
    \small
    \bibliographystyle{ieeenat_fullname}
    \bibliography{main}
}

\maketitlesupplementary
\appendix
\noindent 
The \textit{Appendix} is organized as follows:
\begin{itemize}[leftmargin=*]
    \item \textbf{Section~\ref{supp_sec:algo}:} gives a step-by-step demonstartion of our Nautilus-style Tokenization algorithm.
    \item \textbf{Section~\ref{supp_sec:exp}:} further provides additional experimental results, comparisons, and analyses, including extended visualizations, quantitative evaluations of topological quality, and more in-depth quantitative ablation studies.
    \item \textbf{Section~\ref{supp_sec:discuss}:} provides in-depth discussions on Nautilus, specifically detailing the methodology and experimental comparisons with EdgeRunner.
    \item \textbf{Section~\ref{supp_sec:stat}:} presents the statistical analysis conducted on our training dataset.
    \item \textbf{Section~\ref{supp_sec:user}:} elaborates more details about our user study.
    \item \textbf{Section~\ref{supp_sec:limit}:} discusses the limitation of our approach.
\end{itemize}
We also include a \textit{demonstration video} in our supplementary materials for better visualization.

\begin{figure*}[t]
  \centering
  \includegraphics[width=1.0\linewidth]{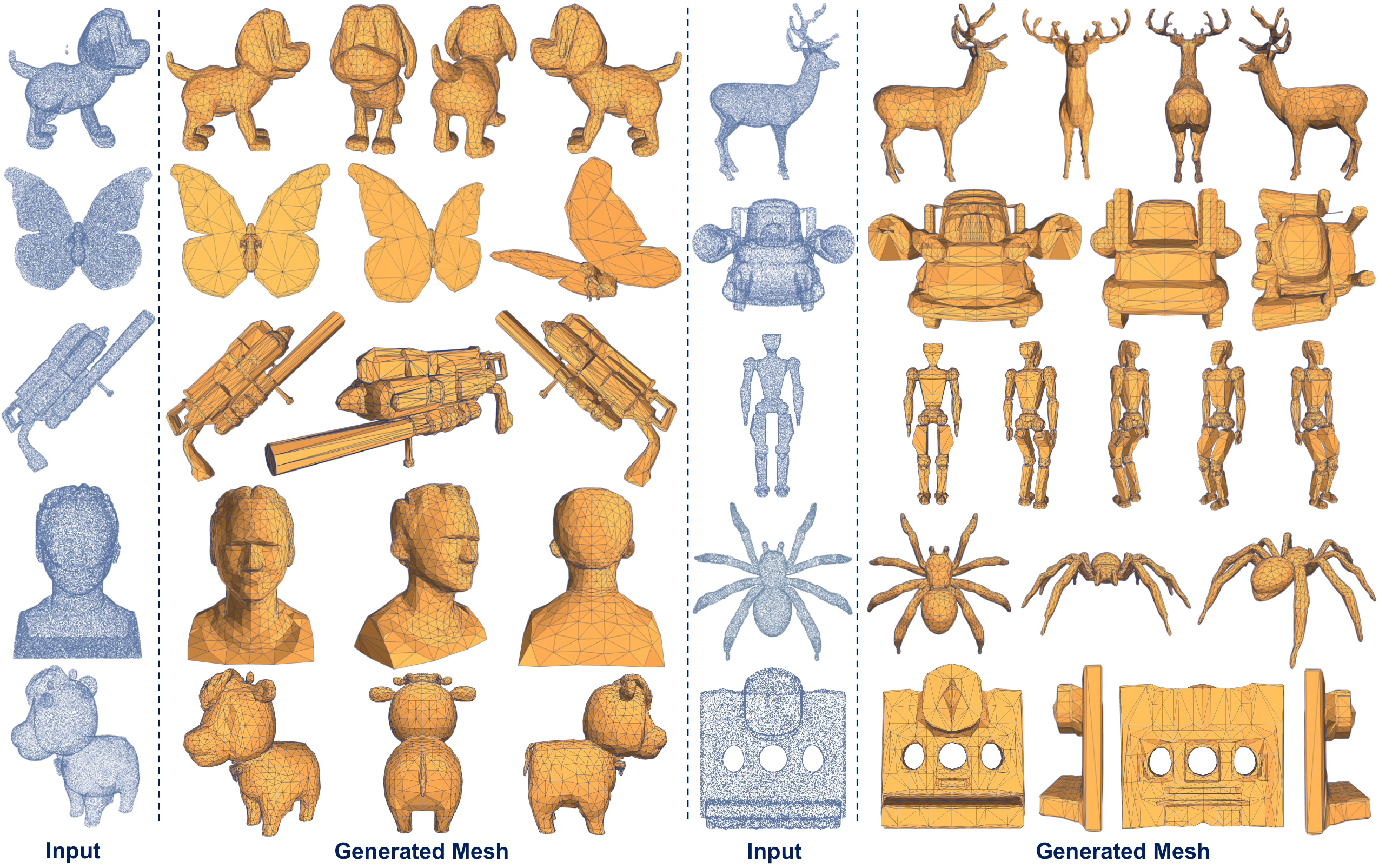}
  \caption{We release additional point cloud conditioned generation results of our Nautilus across an extensive range of object categories.}
  \label{fig:more_pc}
\end{figure*}

\begin{figure*}[t]
  \centering
  \includegraphics[width=1.0\linewidth]{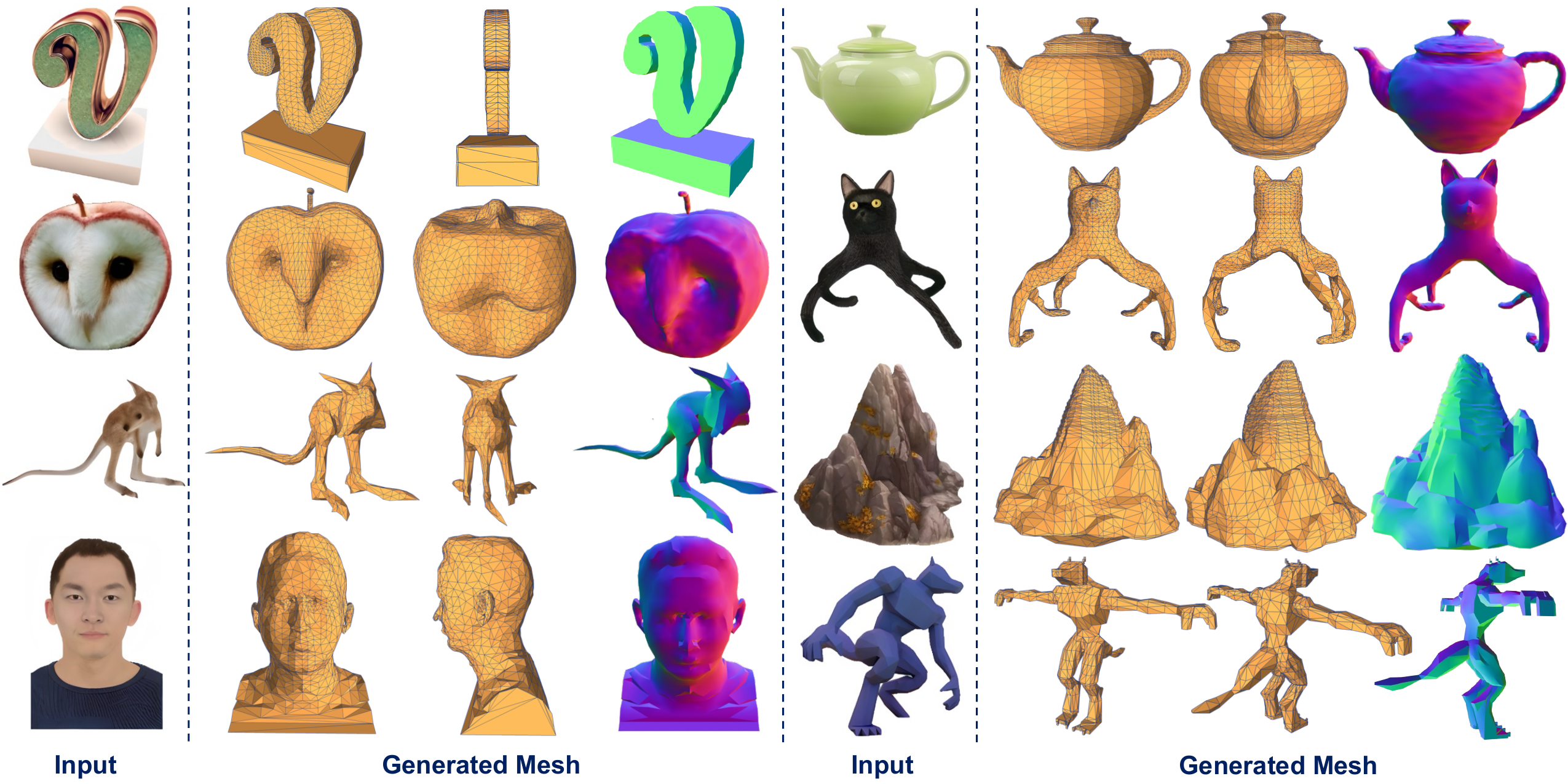}
  \caption{We release additional image conditioned generation results of our Nautilus across an extensive range of object categories. The results are presented alongside corresponding normal maps to highlight the topology.}
  \label{fig:more_img}
\end{figure*}

\input{supp_sec/1_algorithm.tex}    
\input{supp_sec/2_ext_experiment}
\input{supp_sec/3_discussion}
\input{supp_sec/4_furtherstat}
\end{document}

%% file: preamble.tex
\newcommand{\m}{\mathbf{M}}
\newcommand{\res}{\alpha}
\newcommand{\face}{\mathbf{f}}
\newcommand{\vtx}{\mathbf{v}}
\newcommand{\fglb}{f_\text{glb}}
\newcommand{\floc}{f_\text{loc}}
\usepackage{amsmath}
\DeclareMathOperator*{\argmax}{arg\,max}

%% file: sec/0_abstract.tex
\begin{abstract}
Triangle meshes are fundamental to 3D applications.
Current automatic mesh generation methods typically rely on intermediate representations that lack the continuous surface quality inherent to meshes. Converting these representations into meshes produces dense, suboptimal outputs. Although recent autoregressive approaches demonstrate promise in directly modeling mesh vertices and faces, they are constrained by the limitation in face count, scalability, and structural fidelity.
To address these challenges, we propose Nautilus, a locality-aware autoencoder for artist-like mesh generation that leverages the local properties of manifold meshes to achieve structural fidelity and efficient representation. 
Our approach introduces a novel tokenization algorithm that preserves face proximity and compresses sequence length through locally shared vertices and edges, enabling the generation of meshes with an unprecedented scale of up to 5,000 faces. Furthermore, we develop a Dual-stream Point Conditioner that captures fine-grained geometry, ensuring global consistency and local structural fidelity.
Our experiments demonstrate that Nautilus significantly outperforms existing methods in generation quality.
\end{abstract}
\vspace{-3em}

%% file: sec/1_intro.tex
\section{Introduction}
\label{sec:intro}

Triangle meshes, composed of interconnected triangles that approximate arbitrary topology in space, are fundamental to 3D applications~\cite{dai2019scan2mesh, siddiqui2024meshgpt, hong20243dtopia,  chen2020bsp, hanocka2019meshcnn, gong2019spiralnet++, groueix2018papier}. 
Their structure enables efficient modification and rasterization, ensuring compatibility with standard rendering pipelines. 
Currently, creating high-quality meshes primarily relies on skilled 3D artists, making the process labor-intensive and difficult to scale. 
Therefore, automating high-quality mesh generation is essential to meet the demands of modern 3D applications.

However, current automatic approaches mainly adopt intermediate representations such as Voxels~\cite{ren2024xcube,wu2016learning,brock2016generative}, Point cloud~\cite{luo2021diffusion,nichol2022point,zhou20213d}, NeRF~\cite{poole2022dreamfusion,wang2024prolificdreamer,yi2024diffusion, hong2023lrm}, and 3DGS~\cite{tang2025lgm,shen2024gamba,yi2024mvgamba,zhang2024geolrm,wang2025view}, which inherently lack the continuous surface of the meshes. 
Converting these representations into meshes through marching cubes~\cite{lorensen1998marching} or Poisson reconstruction~\cite{kazhdan2006poisson} often yields dense, suboptimal meshes, lacking the conciseness and aesthetic of artist-created models.
Even with post-processing such as remeshing~\cite{attene2006remesh}, the generated meshes remain substantially inferior to artist-created assets, as highlighted by~\cite{chen2024meshanything} and~\cite{siddiqui2024meshgpt}.

Very recently, a line of works has showcased the potential of directly modeling mesh structures~\cite{siddiqui2024meshgpt, chen2024meshanything, chen2024meshxl} in an auto-regressive manner. 
Pioneered by MeshGPT~\cite{siddiqui2024meshgpt}, this paradigm tokenizes mesh faces into sequences and employs a GPT-style decoder to autoregressively generate the sequences, enabling direct capture of vertex-face relationships. MeshAnything~\cite{chen2024meshanything} further enhanced this framework by incorporating point cloud conditioning for global geometric guidance. Through explicit modeling of mesh vertices and faces, these methods generate more concise and structurally refined meshes.

\begin{figure}[t]
  \begin{minipage}[b]{1.0\linewidth}
  \centerline{\includegraphics[width=0.98\linewidth]{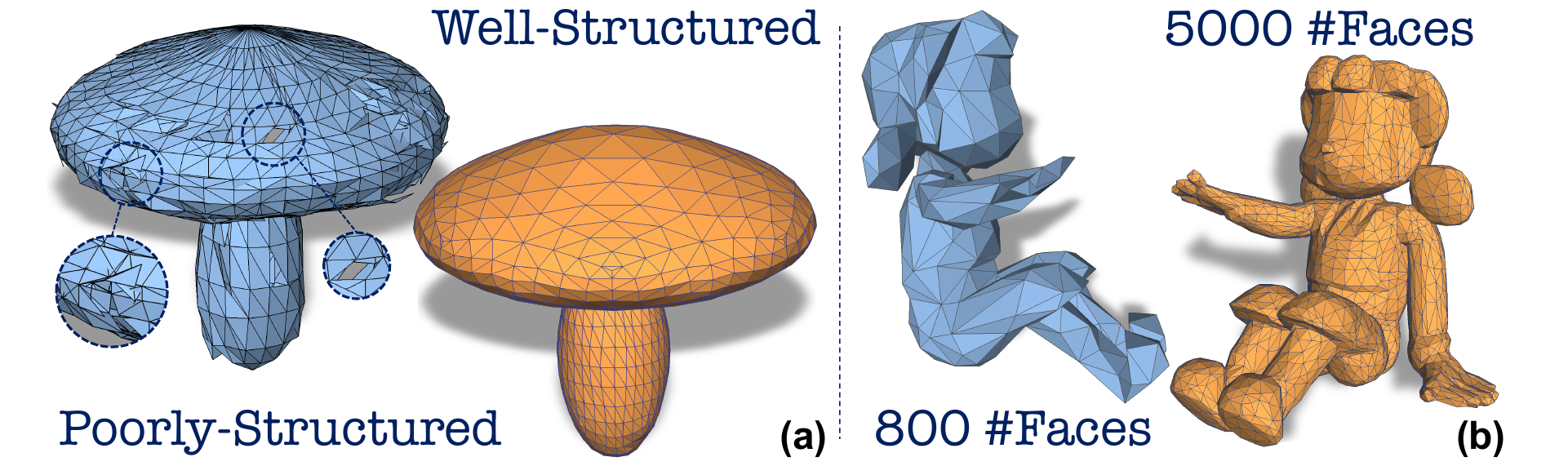}}
  \end{minipage}
  \caption{Fundamental challenges in current methods. \textbf{(a)} Being restricted to generating meshes with fewer faces leads to the loss of fine topological details. \textbf{(b)} Struggling to preserve local structure fidelity causes manifold defects like holes and overlapping faces.}
  \label{fig:challenge}
\end{figure}

Despite these advances, current direct mesh generation still encounters several fundamental challenges:
(1) Existing methods often struggle to maintain local structure fidelity~\cite{weng2024pivotmesh, chen2024meshanything, chen2024meshanythingv2}, as shown in Figure~\ref{fig:challenge}(a), exhibiting manifold defects such as surface holes, overlapping faces, and missing components, especially when dealing with complex topologies.
(2) Moreover, most existing methods are typically limited to modeling meshes with insufficient face number, which significantly constrains their ability to capture the intricate topological details required for industrial applications, as illustrated in Figure~\ref{fig:challenge}(b).

Drawing from artists' creation with precise structure and seamless transitions, we conjecture that the crux arises from inadequate modeling of manifold mesh \textbf{locality}—where adjacent faces share common edges and each face cluster converges at a central vertex.
This locality offers two key inspirations to overcome the aforementioned challenges:
(1) \textbf{Local structure preservation}. 
The locality ensures that each face's geometry is directly constrained by its immediate neighbors.
Thus, prioritizing local dependencies on neighbors and their topology could be crucial for faces to form precise interconnection, achieving manifoldness and structure fidelity rather than merely approximating the shape contour.
(2) \textbf{Efficient mesh representation}. 
Explicitly modeling the locally shared edges and vertices could significantly reduce redundancy in representation and compress the sequence, allowing the modeling of refined meshes with higher face counts and more complex topology.

To this end, we propose \textbf{Nautilus}, a locality-aware autoencoder for scalable artist-like mesh generation. 
First, we design a novel Nautilus-style mesh tokenization algorithm (Sec.~\ref{sec:tokenization}) that preserves the proximity of neighboring faces within the sequence, laying the foundation for modeling local dependencies during generation. In particular, by leveraging the locally shared vertices and edges, it reduces the sequence length to $1/4$ and allows the generation of meshes with up to an unprecedented 5,000 faces.
Second, we introduce a Dual-stream Point Conditioner (Sec.~\ref{sec:conditioner}) that provides multi-scale geometric guidance, ensuring global shape consistency while enhancing local structure fidelity by capturing fine-grained local geometry.
Extensive experiments (Sec.~\ref{sec:exp}) demonstrate that Nautilus significantly outperforms existing methods, delivering high-quality, artist-like mesh generation with intricate topology.

In summary, our key contributions are threefold:
\begin{itemize}[leftmargin=12pt]
\setlength{\parsep}{0pt}
\setlength{\parskip}{0pt}
    \item We pinpoint the neglect of mesh locality as a fundamental limitation in current methods of direct artist-like mesh generation.
    \item We propose \textbf{Nautilus}, which effectively leverages mesh locality through its novel tokenization algorithm along with dual-stream conditioning mechanism.
    \item We demonstrate the scalability of Nautilus through extensive experiments on a large, carefully curated dataset, showing that it achieves high-quality mesh generation with unprecedented topological complexity.
\end{itemize}

\begin{figure*}[t]
  \begin{minipage}[b]{1.0\linewidth}
  \centerline{\includegraphics[width=0.99\linewidth]{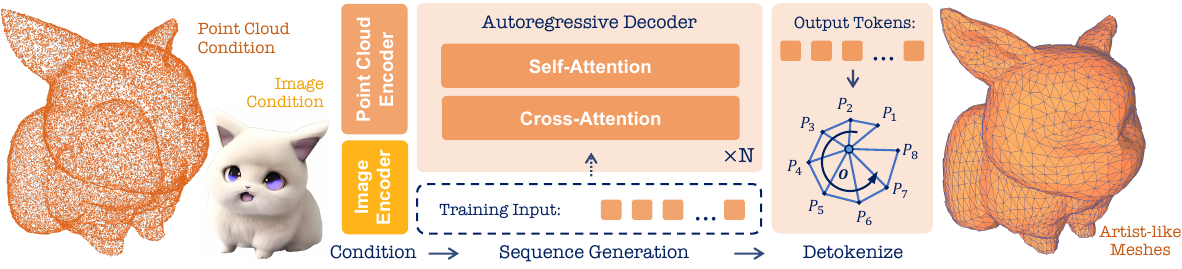}}
  \end{minipage}
  \caption{The overall pipeline of \textbf{Nautilus}. Given a point cloud or image as input, its transformer decoder autoregressively generates mesh sequence tokenized by the Nautilus-style algorithm. Through detokenization, the output tokens are converted to interconnected mesh faces.
  }
  \label{fig:pipeline}
\end{figure*}

%% file: sec/2_related_work.tex
\section{Related Work} 

\noindent \textbf{Mesh Generation with Intermediate Representation.} Most existing 3D generation methods create meshes through intermediate representations, including neural fields~\cite{poole2022dreamfusion,yi2024diffusion,wu2024consistent3d}, 3DGS~\cite{tang2023dreamgaussian,chung2023luciddreamer}, and SDF~\cite{park2019deepsdf,xu2024instantmesh}. Utilizing SDS optimization~\cite{poole2022dreamfusion,lin2023magic3d,wang2024prolificdreamer} with pre-trained 2D models~\cite{rombach2022high,esser2024scaling}, early approaches achieved high quality but required hours of computation per instance. Following the emergence of large-scale 3D datasets, research shifted toward feed-forward 3D models~\cite{hong2023lrm,yi2024mvgamba,tang2025lgm}. Pioneered by the Large Reconstruction Model (LRM), one line of work~\cite{shen2024gamba,zhang2024geolrm} demonstrated that end-to-end training with differentiable rendering and Transformer-based models~\cite{vaswani2017attention} could generate 3D assets within seconds. Another line of work~\cite{zhang2024clay,li2024craftsman,wu2024direct3d} employs 3D latent diffusion with volumetric representation, encoding 3D assets into compressed latent spaces for generation through diffusion steps. However, these methods require decent post-processing~\cite{lorensen1998marching,attene2006remesh} and often produce overly dense or over-smooth meshes.

\noindent \textbf{Direct Mesh Generation.} Compared to methods using intermediate representations, direct mesh generation approaches explicitly model mesh topology, producing more concise and structurally coherent meshes without post-conversion steps.
Pioneering works such as PolyGen~\cite{nash2020polygen} and PolyDiff~\cite{alliegro2023polydiff} showed promise but remained limited to single-category datasets. Recent advances in autoregressive approaches, notably MeshGPT~\cite{siddiqui2024meshgpt}, introduced mesh tokenization with VQ-VAE~\cite{van2017neural} compression for direct mesh generation under topology supervision. Subsequent research has advanced this paradigm by exploring various architectural variants~\cite{chen2024meshxl,weng2024pivotmesh} and extending it to conditional generation tasks~\cite{chen2024meshanything,chen2024meshanythingv2}, particularly point cloud to mesh generation. 
Most recently, EdgeRunner~\cite{tang2024edgerunner} proposes using half-edge algorithm~\cite{rossignac2002edgebreaker} for efficient tokenization.
However, these methods exhibit manifold defects when handling complex topologies and generate excessively long sequences, resulting in a limited number of faces or prohibitively high training costs. In contrast, our Nautilus aims to achieve manifold structure fidelity by a compact, locality-aware mesh tokenization and conditioning mechanism.

%% file: sec/3_method.tex
\begin{figure*}[t]
  \begin{minipage}[b]{1.0\linewidth}
  \centerline{\includegraphics[width=0.98\linewidth]{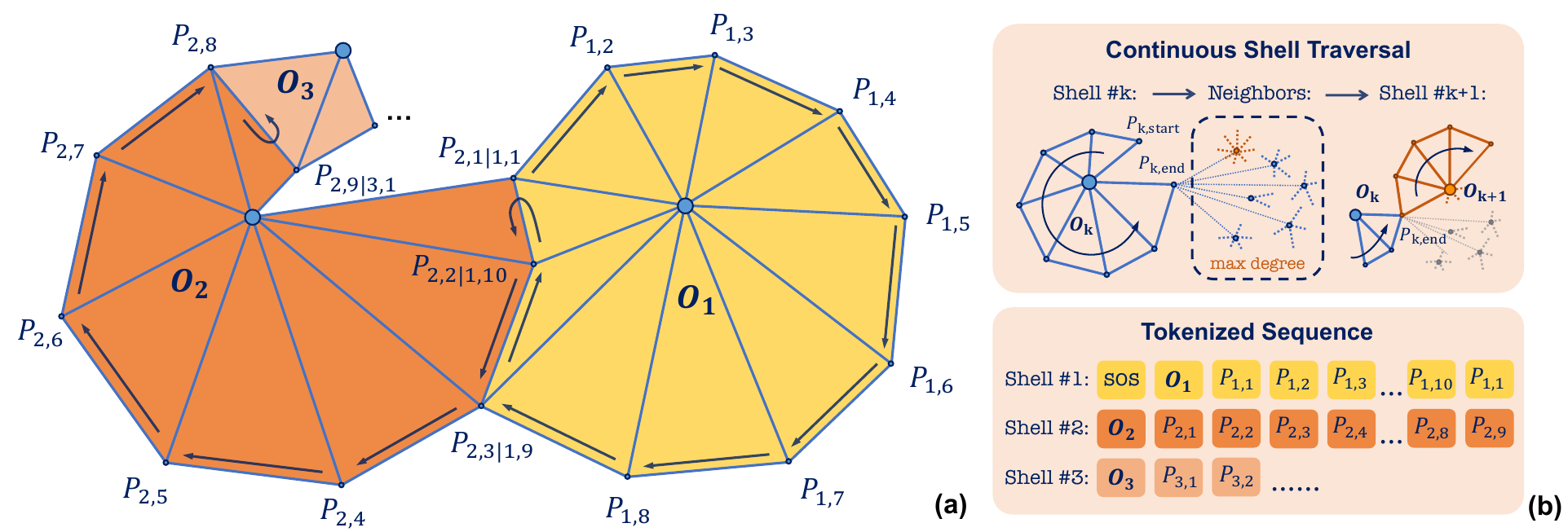}}
  \end{minipage}
  \caption{The illustration of our \textbf{Nautilus-style Mesh Tokenization} algorithm. \textbf{(a)} The shell-structured traversal of mesh faces. \textbf{(b)} \textit{Top}: The continuous shell traversal designed to find the center of the next shell. \textit{Bottom}: The tokenized vertices sequence of the shown example.}
  \label{fig:tokenization}
\end{figure*}

\section{Nautilus}
\label{sec:method}

In this section, we introduce \textbf{Nautilus}, a locality-aware autoencoder for artist-like mesh generation, with the overall pipeline depicted in Figure~\ref{fig:pipeline}.
Initially, we design a novel Nautilus-style algorithm that tokenizes artist-created meshes into sequences (Sec.~\ref{sec:tokenization}), which serves as the training input of our framework.
During generation, Nautilus takes either a point cloud or a single-view image as the condition inputs.
For point cloud conditioning, we propose a dual-stream conditioner (Sec.~\ref{sec:conditioner}) that provides both global and local geometric guidance to the autoregressive decoder, which generates the tokenized sequence given the input condition (Sec.~\ref{sec:decoder}). 
For image conditioning, we integrate a pre-trained image encoder with the decoder to enable image-conditioned generation (Sec.~\ref{sec:image}).
Finally, after detokenization, the token sequences predicted by the decoder are translated into mesh faces, constructing our generated artist-like mesh assets.

\subsection{Nautilus-style Mesh Tokenization}
\label{sec:tokenization}

The initial step of autoregressive mesh generation involves serializing the triangle mesh into tokens~\cite{chen2024meshxl, siddiqui2024meshgpt, chen2024meshanything}. 
A mesh asset $\m = \{\face^i\}_{i=1}^{N}$ consists of $N$ triangle faces $\face^i$, each of which is represented by three vertices, $\{\vtx_1^i, \vtx_2^i, \vtx_3^i\}$. By flattening the discrete 3D coordinates of each vertex, denoted as $\vtx_j^i = \{x_j^i, y_j^i, z_j^i\}$, the mesh asset $\m$ can be serialized into an ordered sequence of coordinates, $S(\m)$, with a total length of $9N$:
\begin{equation}
S(\m) = \{\vtx_1^1, \vtx_2^1, \vtx_3^1, \vtx_1^2, \vtx_2^2, \vtx_3^2, \dots, \vtx_1^N, \vtx_2^N, \vtx_3^N\}
\label{eq:9n}
\end{equation}
However, we argue that such vanilla flattening fails to preserve spatial neighboring in the generated sequence, hindering local dependency modeling. 
Furthermore, its excessive sequence length limits the face count during training, hindering the model from learning complex assets with higher face numbers. 
To overcome these issues, we propose a Nautilus-style tokenization algorithm that maintains local neighbor proximity and reduces sequence length to $1/4$ by leveraging the locally shared vertices and faces.

\noindent
\textbf{Nautilus Shell Representation.} 
In tokenization, we first sort the mesh faces by their coordinates $z$-$y$-$x$ to facilitate efficient serialization. 
As depicted in Figure~\ref{fig:tokenization}(a), we divide the mesh asset into multiple \textbf{shells} for sequential generation. Each shell organizes faces around a central vertex $O$ with an ordered sequence of surrounding vertices $P$, where each face is formed by $O$ and two adjacent $P$ as $\face_{OP_{i}P_{i+1}}$. This organization helps maintain local mesh connectivity by explicitly encoding edge-sharing relationships between adjacent faces.
Based on this shell structure, we compactly represent $N$ faces using a sequence of $(N+2)$ vertices, rather than the $3N$ vertices required in Eq.~\eqref{eq:9n}: 
\begin{equation}
    S(\face_{OP_1P_2}, \face_{OP_2P_3}, \dots) = \{O, P_1, P_2, P_3, \dots\}
\end{equation}
With this representation, the \textit{shell \#1} in Figure~\ref{fig:tokenization}(a) can be serialized to $\{O_1, P_{1,1}, P_{1,2}, \dots, P_{1,10}, P_{1,1}\}$, which preserves the adjacency of neighboring vertices in sequence. 
By explicitly modeling the locally shared vertices and faces, we could add a new face to a shell by extending the sequence with only a single vertex.

\noindent
\textbf{Continuous Shell Traversal.}
After completing the traversal of one shell, we decide the central vertex candidate for the next shell based on two insights:
(1) The vertex should be close to the current shell to facilitate local, short-distance dependencies and promote connectivity between shells.
(2) The vertex should have a relatively high degree and is shared by more faces, which not only improves the compression ratio but also encourages the formation of well-connected local mesh structures.
Therefore, we design the following continuous shell traversal strategy. Specifically, after completing one shell, we list all neighbor vertices of its last traversed vertex and select the neighbor with maximum degree as the center vertex of the next shell~\footnote{As depicted in Figure~\ref{fig:tokenization}(b), the last vertex of the \textit{shell \#k} is located at $P_{k, \text{end}}$, therefore we choose the orange vertex with the highest degree among its neighbors as the next center $O_{k+1}$.}.
Our strategy enhances the traversal continuity and spatial proximity between shells, promoting locally consistent mesh topology.

\noindent
\textbf{Coordinates Compression.}
Building upon the vertex sharing optimization in our Nautilus representation, we exploit spatial quantization and dimensional reduction techniques to achieve more compact mesh encoding. 
Specifically, given the 3D spatial resolution $\res$, we select a proper integer multiplier $\beta$ and uniquely represent the 3D discrete coordinates $(x, y, z) \in \mathbb{R}^\res$ in a 2D space $(u, v)$:
\begin{equation}
x*\res^2 + y*\res + z = u* \beta + v
\end{equation}
By mapping the $x$-$y$-$z$ coordinates to $u$-$v$ space, we could further compress the sequence length.
With a 128 resolution and a multiplier of 2048, we construct a codebook of size 1024 for $u$ and 2048 for $v$.
To identify the start of each shell, we also extend the codebook by 1024 to separately encode the $u$ coordinates of center vertices $O$, thereby distinguishing them from surrounding vertices $P$.
This design avoids the increased sequence length introduced by the special tokens that previous work~\cite{chen2024meshanythingv2, tang2024edgerunner} used for sub-sequence separation, producing our tokenized sequence as:
\begin{equation}
    S(\m) = \{u^\text{O}_{1}, v^\text{O}_{1}, u^\text{P}_{1,1}, v^\text{P}_{1,1}, u^\text{P}_{1,2}, \dots, u^\text{O}_{2}, v^\text{O}_{2}, \dots\}
\end{equation}

In a nutshell, our Nautilus-style tokenization provides a compact serialization to model complex mesh assets with a higher number of faces. 
Meanwhile, it maintains short-distance local dependencies by preserving the proximity of spatially neighboring vertices in the tokenized sequence, enhancing the fidelity of local structure.
Experiments in Sec.~\ref{sec:exp} show that these advantages lead to significant improvements in scalability and generation quality. A detailed step-by-step demonstration of the tokenization process in \textit{Appendix}, further illustrates its design and implementation.

\subsection{Dual-Stream Point Conditioner}
\label{sec:conditioner}

Following~\cite{chen2024meshanything, chen2024meshxl, chen2024meshanythingv2}, our default input condition for Nautilus is point cloud, given its easy accessibility and rich geometric information. We design a dual-stream point conditioner that provides both global and local geometric guidance.

\begin{figure}[t]
  \begin{minipage}[b]{1.0\linewidth}
  \centerline{\includegraphics[width=0.99\linewidth]{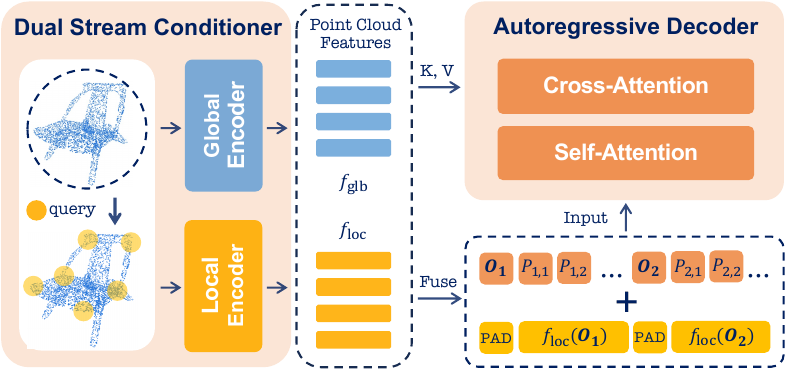}}
  \end{minipage}
  \caption{Illustration of our \textbf{Dual-Stream Point Conditioner}. The Global Encoder extracts global features $\fglb$ from the entire point cloud for cross-attention layers. The Local Encoder captures fine-grained topology information via point convolution, adding local geometric features $\floc$ to each token during generation.}
  \label{fig:pointcloud}
\end{figure}

\noindent
\textbf{Global Point Cloud Encoder.}
In order to capture high-quality global features from the input point cloud, we employ the Michelangelo point cloud encoder~\cite{zhao2024michelangelo} to extract complete geometric information. As illustrated in Figure~\ref{fig:pointcloud}, the point cloud of the entire mesh is encoded into a global latent feature, $\fglb$, which serves as the key and value in the cross-attention layer of the decoder. This global feature encapsulates the overall geometry of the mesh asset, enabling the decoder to grasp the general shape of the input point cloud and autoregressively generate an approximate mesh output that reflects the underlying structure.

\noindent
\textbf{Local Point Cloud Encoder.}
As discussed in Sec.~\ref{sec:intro}, fine-grained local geometric fidelity is crucial for preserving mesh topology and structural coherence.
To this end, we propose a local-aware feature extraction mechanism that aligns with our shell-based generation paradigm, providing precise geometric guidance at the vertex-fan level.
As depicted in Figure~\ref{fig:pointcloud}, we employ a PointConv~\cite{zhang20223dilg} module $\floc(\cdot)$ to capture local geometric information from the point cloud input. 
Moreover, we design a feature injection scheme that augments token embeddings with their local geometric context, inspired by ControlAR~\cite{li2024controlar}. Specifically, for each vertex $P_{k, i}$ in the $k$-th shell, we incorporate the local feature $\floc(O_{k})$ extracted from its central vertex $O_k$ into its tokens feature. For each shell's center vertex, we construct a local neighborhood through KNN sampling of the 100 nearest points, ensuring dense coverage of local surface geometry.
This tight integration with our shell-based tokenization enables progressive incorporation of local geometric constraints during generation, enhancing both topological consistency and structural fidelity.

\subsection{Autoregressive Sequence Decoder}
\label{sec:decoder}

The core generator of our Nautilus is a transformer decoder that autoregressively predicts the probability of the next mesh token based on the preceding.

\noindent
\textbf{Autoregressive Mesh Token Generation.}
After tokenization, the mesh generation task is simply reformulated as an autoregressive sequence prediction problem~\cite{vaswani2017attention,siddiqui2024meshgpt,bai2024meissonic}. 
Using the next-token prediction paradigm, models are trained to predict each discrete coordinate $c \in \{u, v\}$  based on the conditional probability given previously generated tokens. The probability of a mesh \(\m\) can be expressed as:
\begin{equation}
    p(\m) = \prod_{i=1}^{L} p(c_i | c_{<i}), \quad c_i \in \{0, 1, \dots, \res-1\}
\end{equation}
where $\res$ denotes the resolution of the discrete coordinate space, and $L$ is the total length of the generated sequence.

\noindent
\textbf{Training Objective.}
Our decoder is trained under the next token prediction paradigm using cross-entropy loss:
\begin{equation}
    L_\text{CE} = \text{CrossEntropy}(\hat{S}, S(\m)_{>0})
\end{equation}
where $S(\m)_{>0}$ represents the one-hot ground truth token sequence processed by our Nautilus-style Mesh Tokenization, excluding the initial \texttt{[sos]} token, and $\hat{S}$ denotes the predicted classification logits at all positions. 
Consequently, the trained decoder generates tokens that our Nautilus-style tokenization algorithm can accurately detokenize and convert into output meshes.

\subsection{Image Conditioned Generation}
\label{sec:image}
To further expand practical applications, we extend our Nautilus-style tokenization and autoregressive framework to support image-conditioned~\cite{tang2024edgerunner} mesh generation.
We achieve this by leveraging the Michelangelo~\cite{zhao2024michelangelo}, a pre-trained shape generation framework that aligns multiple modalities in a unified feature space.
Specifically, we replace the original decoder in Michelangelo with our mesh decoder, which incorporates Nautilus-style tokenization.
The decoder is trained while keeping the point cloud encoder frozen, enabling mesh generation from the unified feature space.
After training, we integrate Michelangelo's image encoder to enable single-image conditioning.
Using the aligned feature space, our decoder generates high-quality meshes directly from images, as demonstrated by the experimental results in Sec.~\ref{sec:exp}.

%% file: sec/4_experiment.tex
\section{Experiments}
\label{sec:exp}

\begin{figure*}[t]
  \centering
  \includegraphics[width=0.99\linewidth]{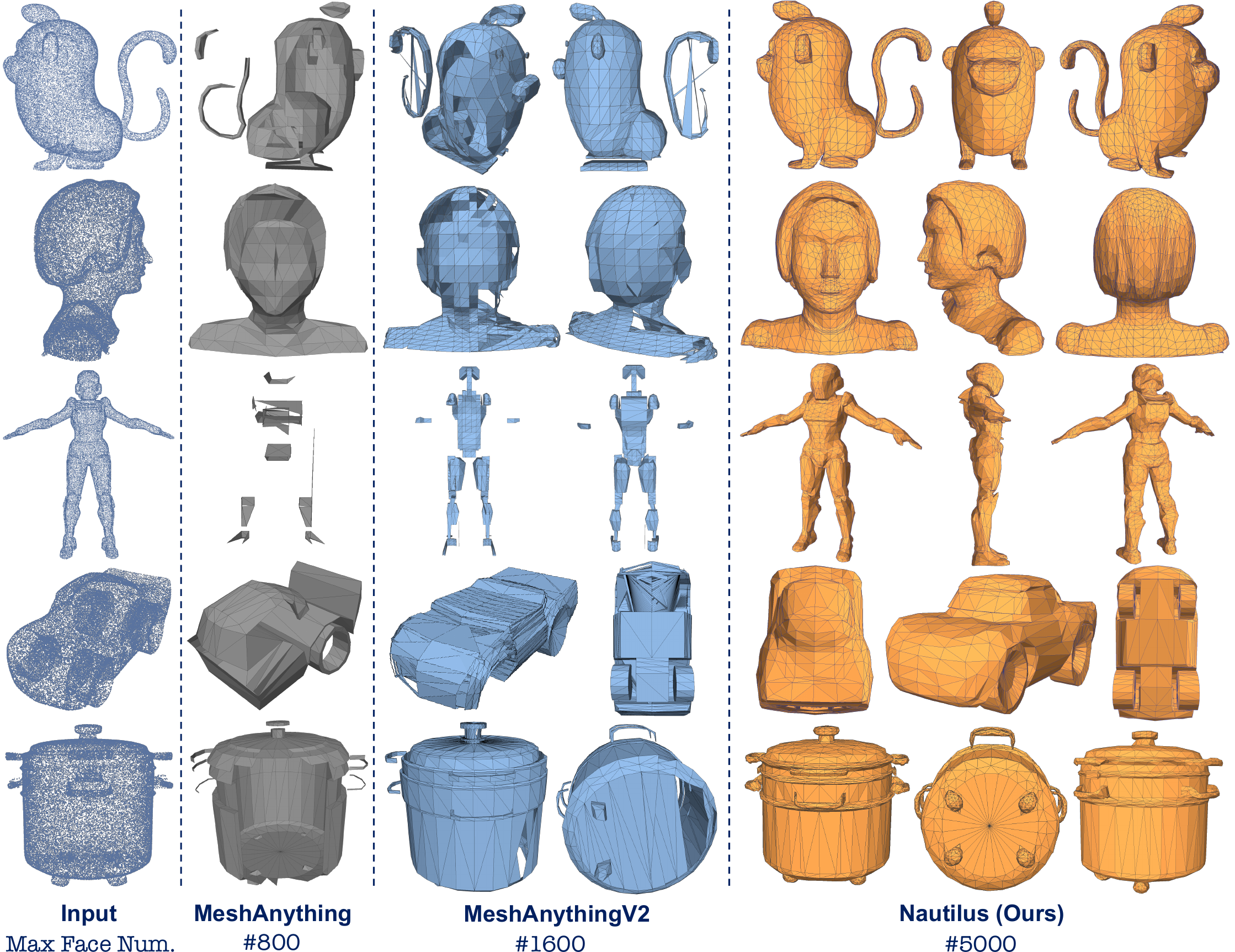}
  \caption{Qualitative comparison on point cloud conditioned generation between Nautilus and latest open-source methods~\cite{chen2024meshanything, chen2024meshanythingv2}.}
  \label{fig:sota}
\end{figure*}

\begin{figure*}[t]
  \centering
  \includegraphics[width=0.99\linewidth]{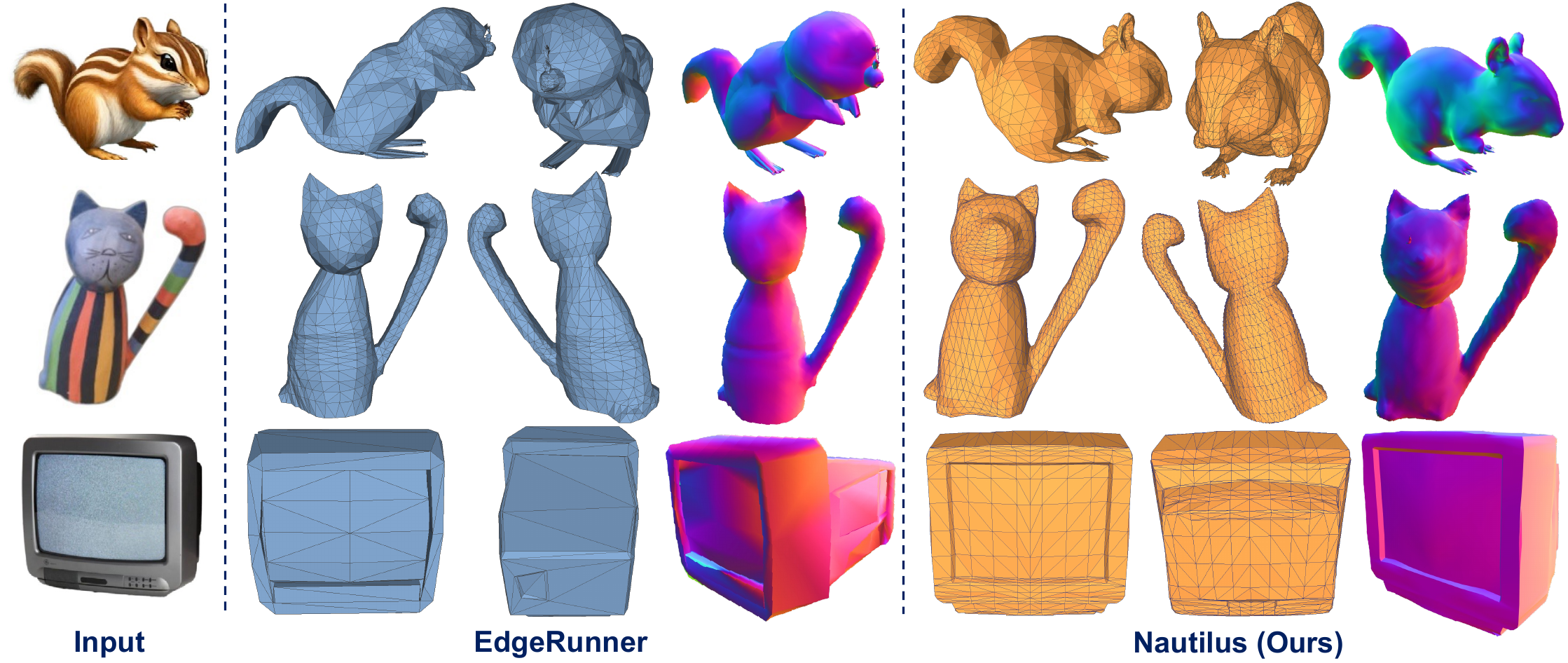}
  \caption{Qualitative comparison on single image conditioned generation between Nautilus and EdgeRunner~\cite{tang2024edgerunner}. The results are presented with corresponding normal maps to highlight topological details.}
  \label{fig:image}
\end{figure*}

\subsection{Implementation Details}
Our Nautilus framework is trained on a curated dataset of 311K high-quality artist-created mesh assets, which is sourced from our manually cleaned Objaverse data and self-collected data, and each containing up to 8,000 faces, with further statistics are provided in \textit{Appendix}.
For each training asset, we sample 4,096 points as input condition.
Nautilus was trained on 8$\times$NVIDIA L40 (48GB) GPUs for 2 weeks, 400K steps, using Adam optimizer with a learning rate of 0.0001 and a batch size of 16. 
The transformer decoder includes 24 layers and contains around 500M parameters.
In inference, we employ probabilistic sampling with a temperature of 0.5 for stability, generating meshes of 5,000 faces in 4 minutes. 
Further statistics of our dataset are in \textit{Appendix}.

\subsection{Qualitative Results}
\label{sec:quali}

\noindent
\textbf{Point Cloud Condition.}
We compare our Nautilus with the latest open-source direct mesh generation methods~\cite{chen2024meshanything, chen2024meshanythingv2} in the point cloud conditioned generation, where EdgeRunner~\cite{tang2024edgerunner} is excluded from this comparison due to its unavailablity of pretrained checkpoints and thus we provide a comparison discussion with EdgeRunner in \textit{Appendix}.
To construct the test set, we use open-source image-to-3D methods~\cite{hunyuan3d22025tencent} to generate dense meshes and directly sample point clouds from them, ensuring that all test samples are entirely unseen during training. 
The selected samples present significant challenges, featuring thin structures, anisotropic faces, and intricate geometric details.

As shown in Figure~\ref{fig:sota}, existing methods struggle with these challenging cases, exhibiting structural defects such as surface holes and missing components, while failing to capture fine topological details. In contrast, Nautilus generates aesthetically appealing, artist-like meshes that faithfully preserve the geometric details of the input. This performance gap stems from their inadequate locality modeling. Current tokenization approaches generate excessive sequence lengths, limiting these methods to training on simple meshes with fewer than 1,600 faces—insufficient for complex test cases. Furthermore, the limited local dependency compromises structural fidelity and increases manifold defects. Nautilus addresses these limitations by leveraging locality in both mesh tokenization and sequence generation, achieving superior quality even for complex topologies, with additional results presented in \textit{Appendix}.

\noindent
\textbf{Single-view Image Condition.}
In Figure~\ref{fig:image}, we present a comparison of meshes generated from single-view images between our Nautilus and EdgeRunner~\cite{tang2024edgerunner}, the only autoregressive method capable of directly conditioning on images~\footnote{Note that our comparison is limited to EdgeRunner's released samples, as their checkpoints are not publicly available.}.
As shown in Figure~\ref{fig:image}, Nautilus generates detailed, manifold meshes with sharp features that accurately match the input conditions, which highlights the superiority of our Nautilus-style tokenization, as further validated in Sec.~\ref{sec:quant}. We present more image-conditioned results and a comparison discussion with EdgeRunner in \textit{Appendix}.

\begin{table}[t]
\centering
\resizebox{.94\linewidth}{!}{
\begin{tabular}{cccc}
\toprule
Metrics        & AMT~\cite{chen2024meshanythingv2} &  EdgeRunner~\cite{tang2024edgerunner} & Ours \\ \hline
Comp. Ratio $\downarrow$& 0.462   &  0.474 &  \textbf{0.275}   \\
Local Ratio $\uparrow$& 0.378    &  0.461 &  \textbf{0.554}   \\ \bottomrule
\end{tabular}
}
\vspace{-0.3em}
\caption{Quantitative comparison with other tokenization algorithms. Our Nautilus-style tokenization achieves more efficient compression while better preserving the local dependency.}
\label{tab:token}
\end{table}

\begin{table}[t]
\centering
\resizebox{.94\linewidth}{!}{
\begin{tabular}{cccc}
\toprule
Methods         & C.Dist. $\downarrow$ & H.Dist. $\downarrow$ & User Study$\uparrow$\\ \hline
MeshAnything~\cite{chen2024meshanything}    & 0.133   & 0.293   & 10.27\%    \\
MeshAnythingV2~\cite{chen2024meshanythingv2}  & 0.106   & 0.248   & 13.17\%    \\
Nautilus (Ours)           & \textbf{0.087}   & \textbf{0.176}   & \textbf{88.68\%}    \\ \bottomrule
\end{tabular}
}
\vspace{-0.3em}
\caption{Quantitative Comparison with latest open-source methods, where our Nautilus framework significantly outperform others in both condition fidelity and visual quality.}
\label{tab:sota}
\end{table}

\subsection{Quantitative Results}
\label{sec:quant}

\noindent
\textbf{Tokenization Algorithm.}
In Table~\ref{tab:token}, we compare our Nautilus-style mesh tokenization algorithm with Adjacent Mesh Tokenization (AMT)~\cite{chen2024meshanythingv2} and EdgeRunner~\cite{tang2024edgerunner} in two aspects:
(1) \textbf{Compression Ratio} measures the reduction in sequence length relative to the original representation $9N$. Higher compression ratios enable modeling of meshes with more faces and better topological complexity.
(2) \textbf{Local Ratio}, a new metric that we introduce, quantifies structural coherence among neighboring tokens. For each vertex, we calculate the proportion of its neighboring vertices in the mesh that appear within 100 preceding positions in the sequence. Higher local ratios indicate better preservation of local dependencies after serialization.
In Table~\ref{tab:token}, our tokenization significantly outperforms other methods in both metrics,
which is attributed to the explicit locality modeling in its design. In contrast, AMT processes mesh faces in an order of elongated strips rather than interconnected local parts, neglecting local dependencies during tokenization, while EdgeRunner relies on the excessive introduction of specialized tokens, leading to suboptimal performance.

\begin{figure*}
  \centering
  \includegraphics[width=0.955\linewidth]{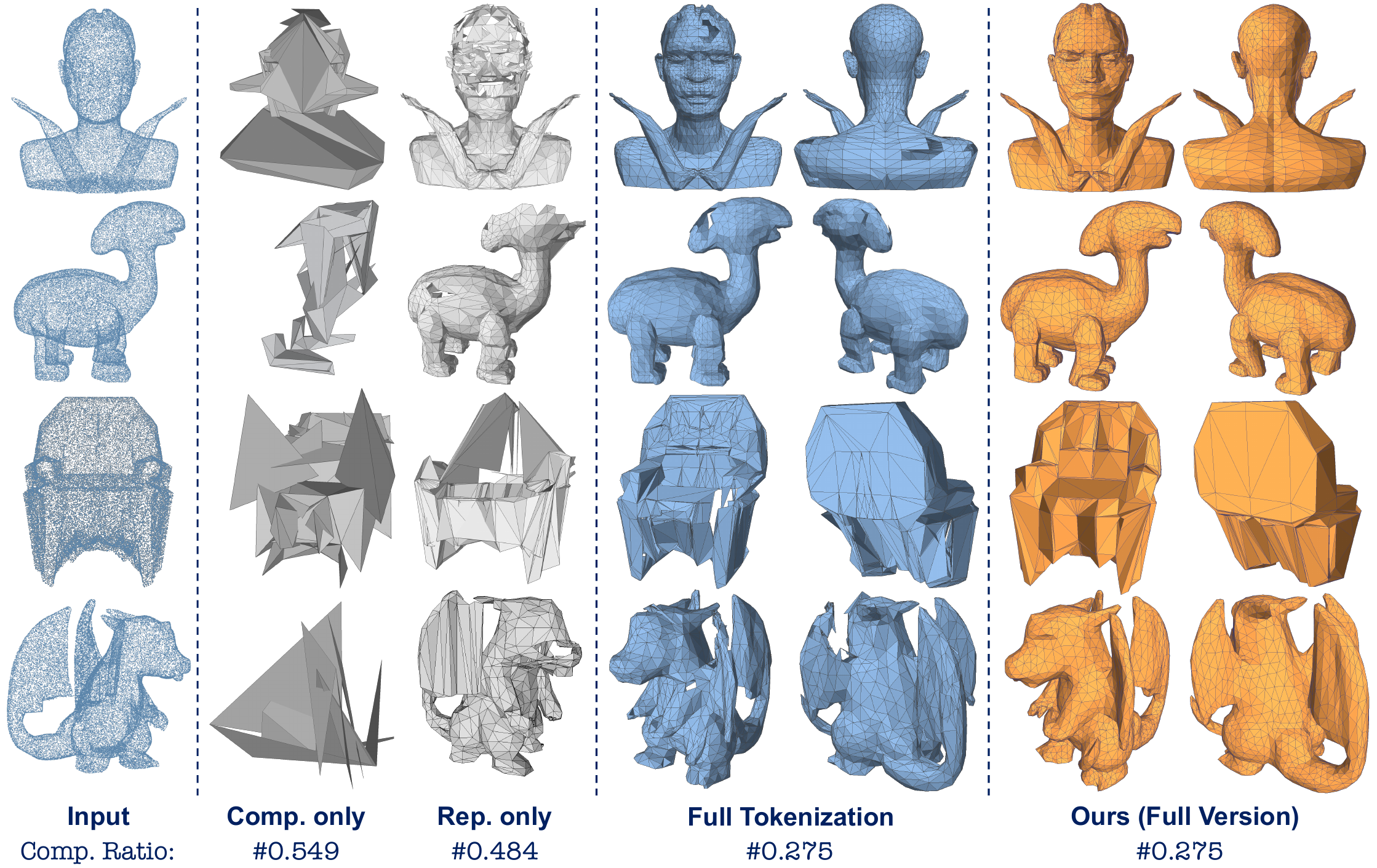}
  \caption{We conduct ablation comparison on four specific configurations: using only the Coordinate Compression (\textit{Comp. only}), using only the Nautilus Shell representation (\textit{Rep. only}), using our \textit{Full Tokenization} but without the local point encoder, and our \textit{Full Version}.}
  \label{fig:ablation}
\end{figure*}

\noindent
\textbf{Generation Performance.}
We compared our point cloud conditioned generation results with MeshAnything~\cite{chen2024meshanything} and MeshAnythingV2~\cite{chen2024meshanythingv2}, on a high-quality test set of 500 samples collected using the same protocol as in our qualitative experiments.
Following the settings of previous work~\cite{tang2024edgerunner}, we evaluate overall reconstruction quality by uniformly sampling 1,024 points from the surfaces of both ground truth and predicted meshes, and computing the Chamfer Distance (C.Dist.) and Hausdorff Distance (H.Dist.) between the sampled point clouds.
Lower distances between point sets indicate higher reconstruction accuracy and better fidelity to the input condition.
In addition, we conducted a user study to assess the proportion of results that users find satisfactory, with further details provided in \textit{Appendix}.
In Table~\ref{tab:sota}, Nautilus significantly outperforms previous approaches, validating its effectiveness that is consistent with the qualitative analysis in Sec.~\ref{sec:quali}. We present further quantitative evaluation on topological fidelity in \textit{Appendix}.

\subsection{Ablation and Discussion}

To validate the effectiveness of our proposed Nautilus-style tokenization and the Dual-Stream Point Conditioner, we perform extensive ablation studies on the tokenization scheme and the local encoder in our conditioner, where a quantitative ablation comparison is presented in \textit{Appendix}.

\noindent
\textbf{Tokenization Scheme.}
We evaluated two configurations to demonstrate the key components of our tokenization design: using only coordinate compression (\textit{Comp. only}) and using only the Nautilus Shell representation (\textit{Rep. only}).
Figure~\ref{fig:ablation} compares these variants with our full version. \textit{Comp. only} achieves a compression ratio of 0.549 and \textit{Rep. only} achieves 0.484, both substantially higher than our full version's 0.275.
In terms of generation quality, \textit{Comp. only} fails to construct the topology completely, while \textit{Rep. only} produces a coarse outline of the target shape. 
In contrast, with the complete tokenization, our full version produces precisely structured results that are consistent with the input condition in geometry.
With the explicit modeling of local neighboring vertices used in our Nautilus Shell representation, both \textit{Rep. only} and our full version demonstrate better performance compared to \textit{Comp. only}.
In addition, as shown in the comparison among the two ablated groups and our full version, insufficient compression excludes complex training samples, resulting in the model’s inability to learn intricate topologies and leading to generation failures.

\noindent
\textbf{Local Point Encoder.}
To evaluate the local encoder's effectiveness in the point conditioner, we compare our full version with a variant that uses full tokenization but excludes the local encoder (\textit{Full Tokenization}). As shown in Figure~\ref{fig:ablation}, the variant without the local encoder exhibits minor manifold defects in complex topological regions, particularly surface holes in local structures. In contrast, our full version, which incorporates the local encoder, successfully resolves these issues and generates precise, compact meshes. The results demonstrate that the local geometric modeling provided by our local encoder yields significant improvements in local details and connectivity, underscoring the importance of enhanced local dependency for generating precise local structures in challenging regions.

%% file: sec/5_conclusion.tex
\vspace{-0.1em}

\section{Conclusion}

In this paper, we present Nautilus, a locality-aware autoencoder designed for scalable high-fidelity mesh generation. By leveraging the locality of manifold meshes, Nautilus introduces a novel tokenization algorithm that preserves local dependency while achieving significant sequence compression, as well as a novel point conditioning mechanism that enhances both global consistency and local structure fidelity.
Our extensive experiments demonstrate that Nautilus outperforms state-of-the-art methods across various metrics, enabling the generation of meshes with unprecedented topological complexity and details. 
These results validate the importance of incorporating locality in tokenization and sequence prediction to achieve superior mesh generation.

%% file: supp_sec/1_algorithm.tex
\section{Tokenization: Step-by-Step Demonstration}
\label{supp_sec:algo}

In our main paper, we design a novel Nautilus-style Tokenization algorithm that preserves local dependency while achieving effective sequence length compression. In Algorithm~\ref{algo:1}, we provide a step-by-step demonstration of our tokenization algorithm, including the initial sorting of vertices and faces, shell construction by vertices traversal, and the coordinates compression of traversed vertices.

\begin{algorithm}[ht]
\scriptsize
\caption{Step-by-Step Pipeline of Nautilus-style Tokenization}
\begin{algorithmic}[1]

\State \textbf{Input:} manifold mesh asset $\mathcal{M} = (\mathcal{V}, \mathcal{F})$ consisting of vertices $\mathcal{V} = \{v_i\}$ and faces $\mathcal{F} = \{f_i\}$.

\State Sort vertices $\mathcal{V}$ by their $z$-$y$-$x$ coordinates.

\State Sort faces $\mathcal{F}$ by the smallest vertex index in each face.

\State Initialize a list of unvisited faces $U_f \gets \mathcal{F}$.

\State Compute the degree $\deg(v_i)$ for all $v_i \in \mathcal{V}$, where $\deg(v_i)$ represents the number of edges connected to vertex $v_i$.

\State Select the highest-degree vertices in the first face $f_1 \in U_f$ as the initial center $O_1$.

\State Initialize the tokenized sequence $S(\mathcal{M}) \gets \varnothing$ and set shell index $k \gets 1$.

\State \textbf{while} $U_f \neq \varnothing$ \textbf{do}:

\State $\;\;\;$ Identify all faces in $U_f$ that contain $O_k$.

    \State $\;\;\;$ Sort these faces in adjacent order, extract their surrounding vertices $\{P_{k, i}\}$.

    \State $\;\;\;$ Extend $S(\mathcal{M}) \gets S(\mathcal{M}) \cup \{O_k, P_{k, 1}, \dots, P_{k, \text{end}}\}$, shell $k$ complete.

    \State $\;\;\;$ Identify all neighbor vertices of $P_{k, \text{end}}$, denoted as $\mathcal{N}(P_{k, \text{end}}) = \{N^i_{k, \text{end}}\}$.

    \State $\;\;\;$ Find the vertex with the maximum degree:
    \[
    N^\text{max}_{k, \text{end}} = \argmax_{N \in \mathcal{N}(P_{k, \text{end}})} \deg(N).
    \]

    \State $\;\;\;$ \textbf{If} $\deg(N^\text{max}_{k, \text{end}}) > 4$ (ensure at least 3 connected faces in the next shell):

    \State $\;\;\;\;\;\;\;\;$ Set $O_{k+1} \gets N^\text{max}_{k, \text{end}}$.

    \State $\;\;\;$ \textbf{Else}:

    \State $\;\;\;\;\;\;\;\;$ Set $O_{k+1}$ as highest-degree vertex in the first remaining face $f_1 \in U_f$.

    \State $\;\;\;$ Update $\deg(v_i)$ for all $v_i \in \mathcal{V}$ to reflect the remaining unvisited edges.

    \State $\;\;\;$ Remove all visited faces from $U_f$, $k \gets k + 1$.

\State Convert the discrete coordinates $(x, y, z) \in \mathbb{R}^{128}$ of vertices in $S(\mathcal{M})$ to $(u, v)$.

\State Flatten the vertices represented in $(u, v)$ coordinates to 1D sequence.

\State Codebook mapping $S(\m)$: $u^\text{P}_{k} = u^\text{P}_{k}$, $v = v + 1024$, $u^\text{O}_{k} = u^\text{O}_{k} + 1024 + 2048$, where $1024$ is the codebook size of $u^\text{P}_{k}$ and $2048$ is the codebook size of $v$.

\State \textbf{Output:} Tokenized sequence $S(\mathcal{M})$.

\end{algorithmic}
\label{algo:1}
\end{algorithm}

%% file: supp_sec/2_ext_experiment.tex
\section{Extensive Experiments and Analysis}
\label{supp_sec:exp}

Our experiments in the main paper comprehensively evaluate the effectiveness of \textbf{Nautilus}. In this section, we provide extensive experiments and analysis to explore behind its efficacy:
\begin{itemize}
\item Sec.~\ref{supp_sec:more_quali}: We present more generation results of Nautilus from both point cloud condition and image condition.
\item Sec.~\ref{supp_sec:topo}: We provide additional quantitative analysis to validate the topological quality of our generated results in comparison with existing methods.
\item Sec.~\ref{supp_sec:quant}: We present an additional quantitative ablation study to complement the qualitative ablation comparisons in the main paper.
\item Sec.~\ref{supp_sec:amt}: We exclusively examine the impact of local dependency on tokenization representation by comparing our Nautilus Shell representation with an upgraded version of the AMT representation that equips our coordinate compression.
\item Sec.~\ref{supp_sec:attn}: We examine the local dependencies modeling by visualizing the attention weights in the generator.
\end{itemize}

\begin{figure*}[ht]
  \centering
  \includegraphics[width=0.97\linewidth]{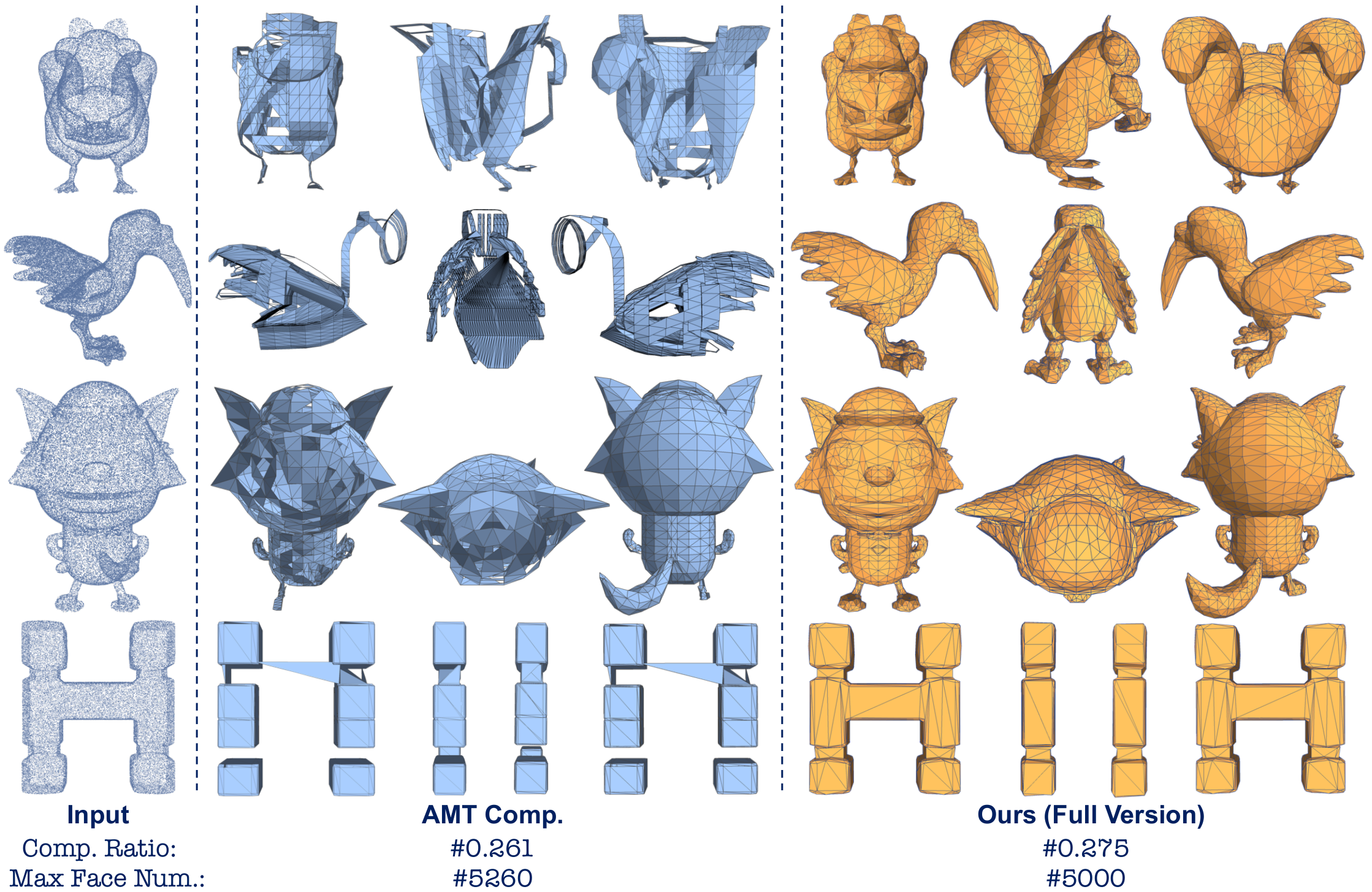}
  \caption{Ablated comparison of generated results between the representation of AMT and our Nautilus Shell representations. To independently analyze the influence of local dependency preservation, both groups are trained on the same dataset, and AMT is improved by incorporating the same coordinate compression as Nautilus to achieve a comparable number of faces, denoted as \textit{AMT Comp.}}
  \label{supp_fig:amt}
\end{figure*}

\subsection{More Qualitative Results}
\label{supp_sec:more_quali}
In Figure~\ref{fig:more_pc} and Figure~\ref{fig:more_img}, we present more generated sample from both point cloud condition and image condition.

\subsection{Quantitative Study on Topological Quality}
\label{supp_sec:topo}

To further validate the performance improvement in locality preservation, we introduce additional quantitative metrics to assess topological quality.
Since no such metric is included in prior works, we adopt two traditional measures and report statistics based on the same 500 point-cloud-conditioned generation results compared in our main paper.
Specifically, we identify surface holes by analyzing boundary edges and use PyMeshLab to detect intersecting faces in the generated results. Additionally, to assess the completeness of generation, we manually inspect for missing parts in the generated meshes.
If no such defects or missing parts are found in a generated mesh asset, we classify it as manifold. Finally, we compute the ratio of meshes exhibiting any of the three defects as well as the proportion of manifold meshes. Notably, a single mesh may exhibit multiple defects simultaneously.

\begin{table}[t]
\centering
\resizebox{1.0\linewidth}{!}{
\begin{tabular}{cccc}
\toprule
Methods         & MeshAnything & MeshAnythingV2 & Nautilus (Ours)\\ \hline
Surf. Holes $\downarrow$     & 45.0\%   & 40.6\%   & \textbf{4.2}\%    \\
Inters. Faces $\downarrow$  & 16.0\%   & 16.8\%   & \textbf{9.6}\%    \\
Miss. Parts $\downarrow$    & 70.4\%   & 58.4\%   & \textbf{8.0\%}    \\ 
Manifold $\uparrow$       & 23.8\%   & 29.0\%   & \textbf{83.6\%}    \\ \bottomrule
\end{tabular}
}
\caption{Quantitative Comparison on Topological Quality with latest open-source methods, where our Nautilus framework significantly outperform others in all topological metrics.}
\label{supp_tab:topo}
\end{table}

We present the results in Table~\ref{supp_tab:topo}, which show that our method achieves a significant improvement across various topological metrics. This is attributed to our approach’s strong locality preservation and high compression rate, enabling the generator model to learn finer local structures from complex topology samples and ultimately produce meshes with higher structural fidelity.

\subsection{Quantitative Ablation Study}
\label{supp_sec:quant}
In our main paper, we conduct a qualitative ablation study on point-cloud-conditioned generation. We here extend the analysis to a quantitative evaluation on the same high-quality test set of 500 samples used in main paper. In Table~\ref{supp_tab:ablation}, we compare four configurations: using only the Coordinate Compression (\textit{Comp. only}), using only the Nautilus Shell Representation (\textit{Rep. only}), using our \textit{Full Tokenization} but without the local encoder, and our \textit{Full Version}.

To assess the generation quality, we follow a similar evaluation protocol: uniformly sampling 1,024 points from the surfaces of both the ground truth and generated meshes. The Chamfer Distance (C.Dist.) and Hausdorff Distance (H.Dist.) are then computed between these sampled point sets to quantify the reconstruction accuracy. Lower values of these metrics indicate a closer alignment between the predicted and ground truth meshes, reflecting higher fidelity to the input conditions.

\begin{table}[h]
\centering
\resizebox{.99\linewidth}{!}{
\begin{tabular}{ccccc}
\toprule
Ablation& Comp. only& Rep. only& Full Tokenization & Full Version\\ \hline
C.Dist. $\downarrow$ & 0.151   & 0.100   & 0.092   & \textbf{0.087} \\
H.Dist. $\downarrow$ & 0.355   & 0.213   & 0.186   & \textbf{0.176} \\ \bottomrule
\end{tabular}
}
\caption{We conduct quantitative ablation comparison on four specific configurations: using only the Coordinate Compression (\textit{Comp. only}), using only the Nautilus Shell Representation (\textit{Rep. only}), using our \textit{Full Tokenization} but without the local encoder, and our \textit{Full Version}.}
\label{supp_tab:ablation}
\end{table}

As shown in Table~\ref{supp_tab:ablation}, our findings are consistent with the qualitative ablation study, confirming that the \textit{Full Version} outperforms all ablated configurations. Notably, the Nautilus Shell representation provides the most substantial improvement (see \textit{Rep. only} vs. \textit{Comp. only}), while Coordinate Compression further contributes to the mesh quality (see \textit{Full Tokenization} vs. \textit{Rep. only}). Additionally, the inclusion of our Local Encoder in the conditioner offers a significant performance boost (see \textit{Full Tokenization} vs. \textit{Full Version}). 
Our findings confirm that achieving both compression and preservation of local dependency, as implemented in our Nautilus, is essential to generate high-quality meshes.

\subsection{Ablation Analysis with AMT Representation}
\label{supp_sec:amt}

To further investigate the influence of \textbf{local dependency preservation} of tokenization algorithms on generation quality, we perform additional experimental comparisons between our Nautilus Shell representation and the AMT in MeshAnythingV2, while excluding the influence of different training data and compression ratio.

Specifically, we apply the same coordinate compression to the AMT representation and train it on the same dataset as Nautilus, enabling it to model more than 5,000 faces. In Figure~\ref{supp_fig:amt}, we denote the improved AMT version as \textit{AMT Comp.}, comparing it with our full version on challenging samples with complex topology and rich geometric details. Despite achieving a compression ratio comparable to that of Nautilus, \textit{AMT Comp.} still exhibits severe manifold defects in its generation results, particularly \textbf{isolated strips} consisting of single-line faces.

This issue arises from the lack of local dependency preservation in AMT’s traversal. Unlike the part-by-part approach taken by human artists, AMT processes mesh faces in elongated strips, making it extremely challenging for the model to precisely connect these strips into manifold meshes. 
Without preserving local dependency, for each face in the sequential generation, its neighboring faces in the mesh are often located far apart in the sequence.
This requires the decoder to build highly precise features to globally align and gather information from these distant neighbors. As these neighbors are essential for determining the shape of the current face according to mesh locality, failing to aggregate their information leads to poor interconnections and manifold defects.

In contrast, Nautilus preserves short-distance, local dependency, maintaining the proximity of spatially neighboring vertices in the tokenized sequence. During the generation of each face, this local dependency introduces an inductive bias, facilitating the decoder to simply gather information from nearby positions in the sequence. This approach enables Nautilus to achieve significantly better local interconnection and structural fidelity. Our results highlight the importance of preserving local dependency in tokenization algorithms for generating high-quality meshes. Additional visualization analysis is provided in Sec.~\ref{supp_sec:attn}.

\subsection{Visualized Analysis on Local Dependency}
\label{supp_sec:attn}

In Supp. Sec.~\ref{supp_sec:amt}, we conducted additional ablation comparisons to demonstrate the significance of preserving local dependency within tokenization algorithms.
In this section, we present a visualized analysis to directly examine the modeling of local dependency. Specifically, we compare the averaged attention maps of our Nautilus and that of the ablated group \textit{AMT Comp.} designed in Supp. Sec.~\ref{supp_sec:amt}. For each model, we compute the averaged causal attention map across 50 samples and all 24 self-attention layers in the transformer decoder, in order to evaluate the models’ capacity to model local dependency.

\begin{figure}[t]
  \centering
  \includegraphics[width=0.99\linewidth]{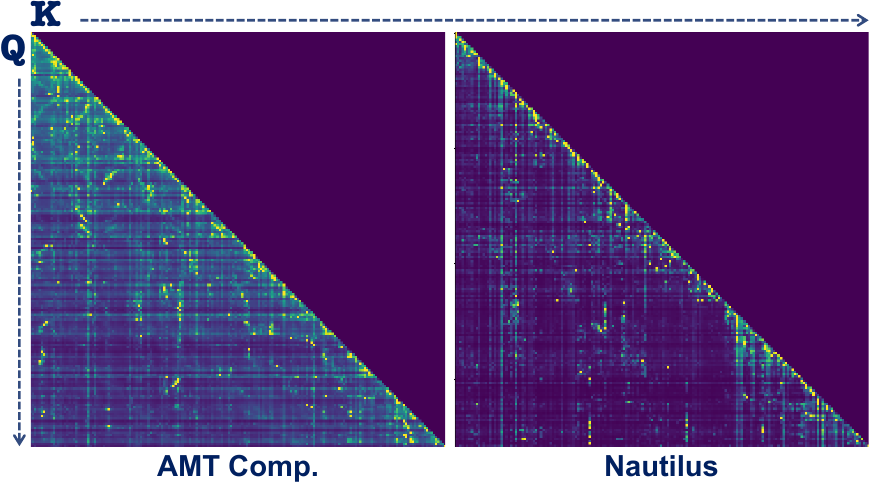}
  \caption{Comparison of averaged Attention Maps between \textit{AMT Comp.} and Nautilus. The attention maps are averaged over 50 samples and all 24 self-attention layers in the decoder. Each row corresponds to a query (Q) position, and each column corresponds to a key (K) position. Bright colors indicate stronger attention weights directed from a Q position to a K position.}
  \label{supp_fig:attn}
\end{figure}

We present our findings in Figure\ref{supp_fig:attn}, where each row represents a query (Q) position and each column represents a key (K) position. Bright colors indicate stronger attention weights directed from a Q position to a K position. From Figure\ref{supp_fig:attn}, it is evident that Nautilus exhibits highly concentrated bright regions close to the diagonal. In other words, for each query position, the attention is predominantly focused on a few nearby positions within the sequence.
In contrast, the attention map of \textit{AMT Comp.} shows more dispersed bright regions, suggesting that attention is distributed to positions farther away in the sequence compared to Nautilus. This visualization clearly demonstrates the impact of local dependency preservation on the performance of decoder.
By effectively maintaining the proximity of spatially neighboring vertices in the tokenized sequence, the transformer decoder in Nautilus gathers local topology information more efficiently from nearby positions within the sequence, which contributes to its superior quality.

%% file: supp_sec/3_discussion.tex
\section{Further Discussions}
\label{supp_sec:discuss}

\subsection{Comparison with EdgeRunner}
\label{supp_sec:comp_er}
Due to the unavailability of checkpoints and the absence of normals in their point cloud demo, we are unable to conduct a direct comparison with EdgeRunner under point cloud conditions. Therefore, in this section, we provide a methodological and experimental comparison to highlight the differences between our approach and EdgeRunner.

\subsubsection{Methodological Comparison with EdgeRunner}
In terms of methodology, both Nautilus and EdgeRunner follow the auto-regressive mesh generation paradigm pioneered by MeshGPT, yet they share few similarities beyond that. While EdgeRunner propose using the half-edge representation, our Nautilus approaches focus on preserving locality structure and exploiting it for more advanced tokenization compression.

\noindent
\textbf{Tokenization Algorithm.}
EdgeRunner is built on the half-edge algorithm and relies on an excessive number of separation tokens for faithful traversal, resulting in a suboptimal compression ratio and locality preservation (as shown in Table 1 of our main paper). Limited compression capability constrains the maximum number of faces and increases training costs, while weaker locality preservation makes the generated results more susceptible to local structural defects and the loss of fine geometric details.
On the other hand, Nautilus leverages shared vertices and edges within mesh locality and proposes Nautilus-style representation with continuous shell traverse. 
This design naturally aligns with the mesh creation flow and eliminates the need for excessive separation tokens, allowing it to surpass EdgeRunner in both locality preservation and sequence compactness. As a result, our approach achieves lower training costs, a higher maximum face count, and finer local topological quality, which is demonstrated in Figure 7 in main paper.

\noindent
\textbf{Conditioning Mechanism.}
For point cloud conditioning, EdgeRunner employs a global encoder similar to previous methods, while we introduce a novel PointConv-based Local Encoder along with a seamless injection scheme that integrates local features into our Nautilus-shell representation, effectively enhancing local dependency. 

\subsubsection{Experimental Comparison with EdgeRunner}

\noindent
\textbf{Qualitative and Quantitative Comparison.}
In our main paper, we made a significant effort to conduct performance comparisons as comprehensively as possible.
Our quantitative analysis demonstrates that our tokenization algorithm surpasses EdgeRunner in both compactness and locality preservation, with a compression capability nearly twice that of EdgeRunner.
Furthermore, as shown in Figure 7, Nautilus achieves superior qualitative results, significantly outperforming EdgeRunner in image-conditioned generation by producing more detailed outputs. This improvement stems from our advanced tokenization scheme, which not only achieves a higher compression ratio but also better preserves locality.

\noindent
\textbf{Backbone and Training Cost.}
Regarding the backbone size and training step of the generator, our method employs a 24-layer transformer with 1024 hidden dimensions, which is \textbf{more lightweight} than EdgeRunner that uses a 24-layer transformer with 1536 dimensions. In terms of training cost, our method was trained on 8$\times$L40 (48GB) GPUs for 2 weeks, whereas EdgeRunner used 64$\times$A100 (80GB) GPUs for 1 week. Thus, our total GPU hours are \textbf{less than 1/4} of those used by EdgeRunner. Besides, we employs a training-free Michelangelo conditioner for image input, whereas EdgeRunner used 16$\times$A800 (40GB) GPUs for 1 week on training image conditioner.

\noindent
\textbf{Maximum Face Ability.}
Due to our GPU resource limitations, we train Nautilus with a maximum of \textbf{12,000} tokens, whereas EdgeRunner, utilizing 64$\times$A100 (80GB) GPUs, trains with about \textbf{17,200} tokens, which can be obtained from their maximum face number of \textbf{4,000} and compression ratio of 0.474. Theoretically, given the significant improvement in compression ratio, our tokenization supports the training and generation of \textbf{7,000} faces assets with the same 17,200 tokens on comparable GPU resources -- \textbf{nearly twice} that of EdgeRunner. In the future, we are planning to extend our training data to cleaned higher face number mesh assets and further scale up our framework.

In summary, based on novel methodologies, Nautilus focuses on improving local structure modeling and representation compactness using mesh locality, achieving significantly higher generation quality with more faces and far less training cost comparing with EdgeRunner.

\subsection{Impact of Sequence Compression}
\label{supp_sec:length}

In this section, we provide a comprehensive discussion on how compression ratio directly impacts generation quality during both training and inference stages.

\noindent
\textbf{Effect in Training.}
During training, artist-created mesh assets in the dataset are converted into sequences through tokenization. Due to GPU memory limits, the maximum sequence length of a 500M parameter transformer decoder in autoregressive mesh generation is typically around 20,000. For efficient training, sequence lengths are generally constrained around 12,000.
Within this limit, tokenization algorithms with higher compression ratios enable the inclusion of more complex, high-face-count mesh assets. Learning from such assets enables the model to handle complex geometries under challenging conditions, mitigating failures such as missing components, large surface holes, and other structural defects.
With Nautilus achieving an unprecedented compression ratio of 0.275, it allows the inclusion of mesh assets containing over 5,000 faces in 12,000 tokens. This compression facilitates the model’s ability to learn from high-quality assets, enabling the generation of unprecedented topological complexity.

\noindent
\textbf{Effect in Inference.}
The inherent gap between the prediction of the next token during training and autoregressive inference during inference amplifies the cumulative errors as the length of the sequence increases.
Our empirical analysis reveals that, even with well-optimized models, the validation loss at the 15,000-th position is approximately three times higher than at the 5,000-th position.  
Therefore, tokenization algorithms with lower compression ratios further aggravate this issue by requiring longer sequences to represent the same number of faces, increasing the likelihood of defects. This observation aligns with our ablation study, where the groups using only Nautilus Shell Representation (\textit{Rep. Only}) demonstrates inferior generation quality compared to the \textit{Full Tokenization}, which is attributed to its inferior compression ability.

\begin{figure}[t]
  \centering
  \includegraphics[width=0.99\linewidth]{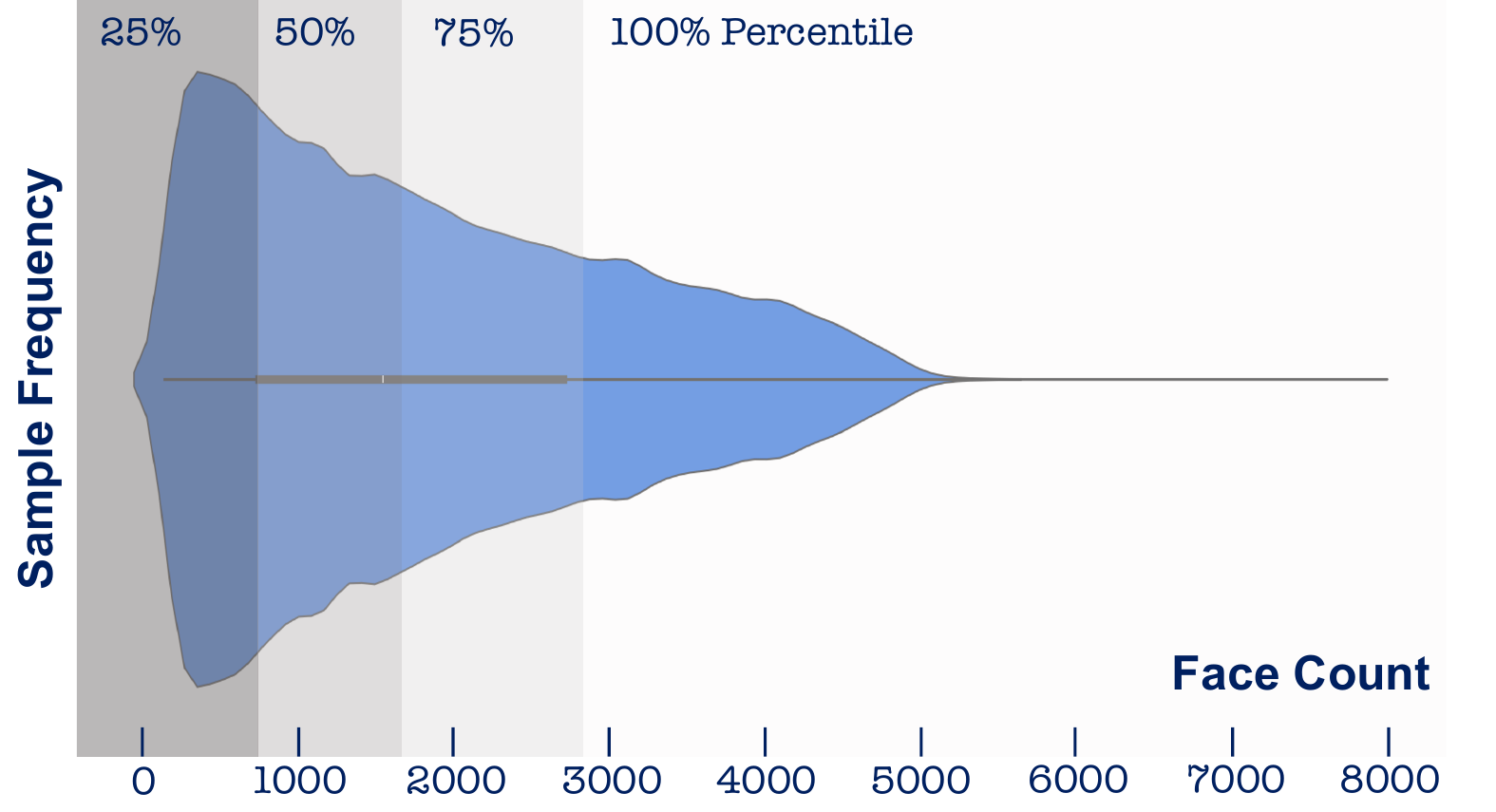}
  \caption{Statistics on the samples' face counts in our training set, including 311K high-quality mesh assets with up to 8,000 faces.}
  \label{supp_fig:stat}
\end{figure}

%% file: supp_sec/4_furtherstat.tex
\begin{figure*}[t]
  \centering
  \includegraphics[width=0.99\linewidth]{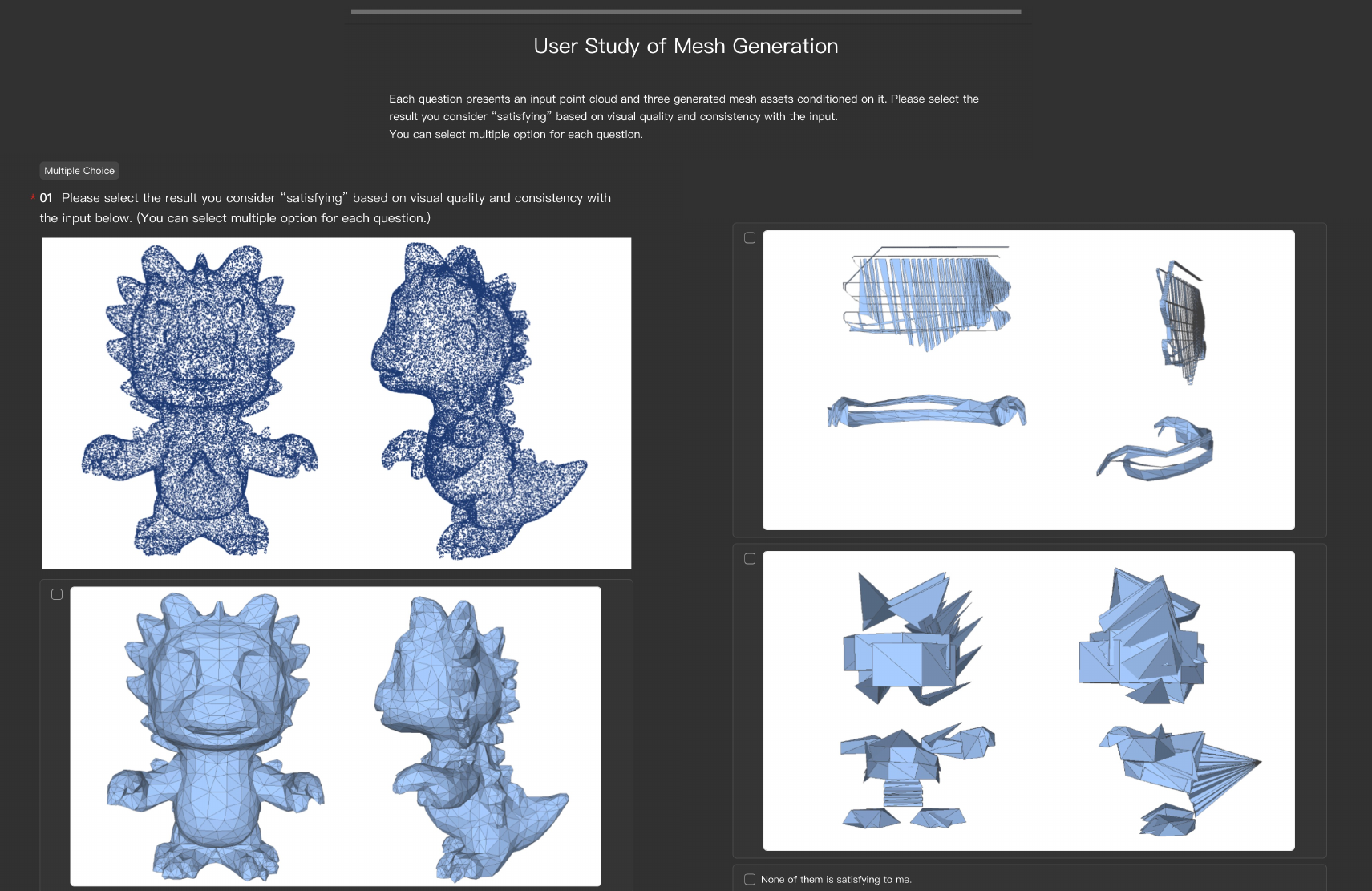}
  \caption{The Interface of Our User Study. Given the rendered images of the input condition and the generated meshes from each compared method, users are asked to indicate whether they are satisfied with each result. We calculate the average satisfaction rate for each method.}
  \label{supp_fig:userstudy}
\end{figure*}

\section{Training Data Statistics}
\label{supp_sec:stat}

We conducted a statistical analysis of the face counts in our training dataset, with the results shown in Figure~\ref{supp_fig:stat}.
Our training dataset consists of 311K high-quality mesh assets that can be tokenized into 12,000-length sequences using our Nautilus-style Tokenization Algorithm.
As illustrated in the figure, about 50\% of the samples exceed the maximum face count of 1,600 supported by previous methods and even include a small subset of extremely complex samples with 5,000 to 8,000 faces. 
This demonstrates that our tokenization method, while typically supporting meshes with up to 5,000 faces, is capable of achieving an even higher level of compression in certain cases, enabling the tokenization of meshes with up to 8,000 faces.
Leveraging this extensive and carefully curated training set, which covers a wide range of face counts, Nautilus not only ensures basic geometric modeling but also learns the ability to effectively handle complex topologies.

\section{User Study Details}
\label{supp_sec:user}
In this section, we provide additional details of our User Study.
As shown in Figure~\ref{supp_fig:userstudy}, each question in our study includes the rendering visualizations of the input point cloud and the output meshes generated by the compared methods. To ensure unbiased feedback, all options are presented anonymously, and their order is shuffled for every question. 
Specifically, users are asked to evaluate each result by indicating whether they are satisfied with it. 
For each method, we collect the responses and calculate the average satisfaction rates across all questions. In total, our User Study consists of 20 questions, featuring 20 input point clouds and 60 corresponding mesh assets generated from the three comparing methods. In total, the survey received 38 responses from participants with diverse backgrounds.
This carefully designed questionnaire, combined with a broad participant base, ensures the objectivity and reliability of the results.

\section{Limitations}
\label{supp_sec:limit}
Through the locality-aware mesh tokenization and local-dependency enhancement mechanisms, our method effectively improves the manifoldness of local mesh structures and topological quality.
However, it does not strictly guarantee the generation of fully manifold meshes. Specifically, while our approach significantly reduces surface holes, the improvement in reducing intersecting faces is relatively relatively less pronounced. In future work, we plan to further address this issue by introducing more explicit structural constraints.